\documentclass[letterpaper, 10 pt, conference]{ieeeconf}  

\IEEEoverridecommandlockouts                              

\usepackage[english]{babel}
\usepackage[autostyle]{csquotes} 
\usepackage{authblk}
\MakeOuterQuote{"}

\usepackage{graphicx}
\usepackage{gensymb}

\makeatletter
\let\NAT@parse\undefined
\makeatother

\usepackage[pagebackref=true,breaklinks=true,letterpaper=true,colorlinks,urlcolor=black,bookmarks=false]{hyperref}

\title{\LARGE \bf
Learning from Maps: Visual Common Sense for Autonomous Driving
}

\author{Ari Seff \\ \small{\tt aseff@princeton.edu} \and Jianxiong Xiao \\  \small{\tt profx@autox.ai} 
\thanks{Ari Seff is with the Department of Computer Science, Princeton University, Princeton, New Jersey.
}
\thanks{Jianxiong Xiao is with AutoX, Inc., San Jose, California.
}}

\begin{document}

\maketitle
\thispagestyle{empty}
\pagestyle{empty}

\begin{abstract}

Today's autonomous vehicles rely extensively on high-definition 3D maps to navigate the environment. While this approach works well when these maps are completely up-to-date, safe autonomous vehicles must be able to corroborate the map's information via a real time sensor-based system. Our goal in this work is to develop a model for road layout inference given imagery from on-board cameras, without any reliance on high-definition maps. However, no sufficient dataset for training such a model exists. Here, we leverage the availability of standard navigation maps and corresponding street view images to construct an automatically labeled, large-scale dataset for this complex scene understanding problem. By matching road vectors and metadata from navigation maps with Google Street View images, we can assign ground truth road layout attributes (e.g., distance to an intersection, one-way vs. two-way street) to the images. We then train deep convolutional networks to predict these road layout attributes given a single monocular RGB image. Experimental evaluation demonstrates that our model learns to correctly infer the road attributes using only panoramas captured by car-mounted cameras as input. Additionally, our results indicate that this method may be suitable to the novel application of recommending safety improvements to infrastructure (e.g., suggesting an alternative speed limit for a street). 

\end{abstract}

\section{INTRODUCTION}

The effort to develop autonomously driven cars has seen tremendous progress in recent years. Several automobile manufacturers and technology companies have showcased cars that can now drive on their own in a variety of settings, including highways and limited urban environments. However, for precise car localization and access to road layout information, these systems rely on ultra high-definition (HD) maps that must be pre-scanned prior to deployment of a driverless vehicle. While this is an adequate approach when up-to-date maps and strong GPS signals (to initialize the car's localization) are available, such a dependency leaves autonomous vehicles vulnerable to malfunctioning in their absence. These high-definition maps can become inaccurate as construction and road modifications occur. Additionally, skyscrapers and other infrastructure often impede GPS signals, limiting the viability of self-localization in urban settings \cite{intersection}.

\begin{figure}
\centering 
\includegraphics[width=3.45in]{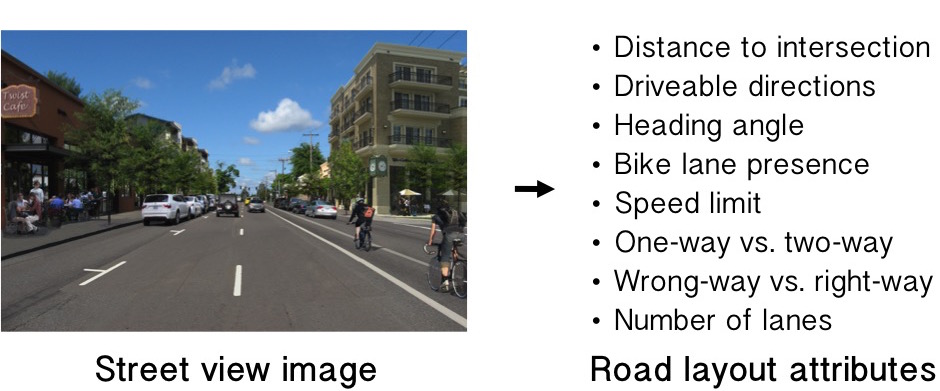}
\vspace{-3ex}
\caption{Given a street view image, our model learns to estimate a set of driving-relevant road layout attributes. The ground truth attribute labels for model training are automatically extracted from standard navigation maps.}
\label{fig:teaser}
\end{figure}

In this work, our goal is to develop a model that estimates a set of road layout attributes given a single RGB street view image captured by a car's on-board camera (Figure \ref{fig:teaser}). Included in this attribute set are driveable path headings, distances to intersections, traffic directionality (one-way vs. two-way), and several others (see Sec. \ref{sec:targets} for a complete list with descriptions). Such a system can serve as a backup for safety and reliability, corroborating the HD map in real time. In any technology where fatal consequences result from incorrect functioning, this minimal level of redundancy is essential. 
Additionally, if this map-like information can be inferred on the fly, autonomous cars may eventually be unrestricted, able to venture beyond the boundaries of high-definition maps.

However, road layout inference is quite a challenging task. There exists substantial variability in types of intersections, street/lane width, and road appearance, especially in urban environments. Additionally, because of infrastructure, vegetation, and other vehicles, portions of the road and upcoming intersections are often occluded. Having a very large amount of training data is absolutely crucial to developing a well-generalized model.

Here, we leverage the availability of street view image databases and standard navigation maps to construct a large-scale dataset\footnote{All code, data, and pre-trained models are available at \url{http://www.cs.princeton.edu/~aseff/mapnet}.} for this complex scene understanding task. By harnessing this source of images and corresponding map information, we bypass both manual image capturing (e.g., with a data-collection vehicle \cite{deep_highway}) and intensive manual annotation of ground truth labels. 

To construct this dataset, we gather one million Google Street View (GSV) images to serve as input instances to our model. These images are similar to those acquired by today's highly abundant and inexpensive car-mounted cameras, albeit they are panoramas with a slightly higher mount point.
In order to label the images with a ground truth description of the driving scene, we extract information from OpenStreetMap\footnote{\url{www.openstreetmap.org}} (OSM). 
As an open-source map, OSM stores road vectors and descriptive metadata that are freely available for download.

\begin{figure}
\centering 
\includegraphics[width=3.45in]{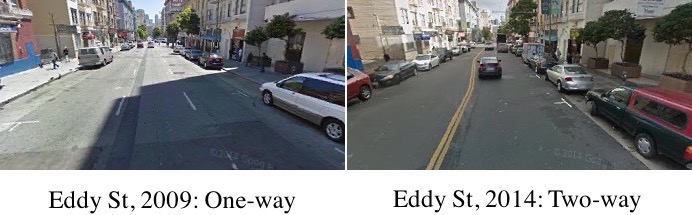}
\caption{Two Google Street View images are shown from the same segment of Eddy St., San Francisco in 2009 (left) and 2014 (right). Our trained model can handle infrastructure modifications over time without re-annotation, correctly identifying the scene on the left as a one-way street and the right as a two-way street.}
\label{fig:twoway_roadmod}
\end{figure}

Given this high volume of annotated data, the road attribute inference problem becomes amenable to a deep learning approach. 
Here, we exploit the recent success of convolutional neural networks (ConvNets) across a range of computer vision tasks and train models for this specific problem. Standard ConvNet architectures equipped for classification or regression allow for estimating both categorical and numerical road attributes. Importantly, the trained models can cope with modifications to infrastructure over time without any re-annotation (Figure \ref{fig:twoway_roadmod}).

The contributions of this work are three-fold: (1) We construct an automatically-labeled image dataset for road layout inference larger than any previously available. (2) We demonstrate that convolutional networks may be trained on this self-supervised data to accurately predict a set of road scene attributes given a single RGB panorama, without any reliance on HD maps. (3) We further find that such models may also be employed to recommend improvements to city infrastructure (e.g., recommending an alternative speed limit for a road) if the ground truth attribute set for a road does not match the set indicated by its visual appearance. 

\section{Related Work}
Here we review work spanning road layout inference, deep learning as applied to autonomous driving, and the use of OSM/GSV data for developing other applications.

\textbf{Estimating road layout from sensor input:}
In \cite{intersection}, Geiger et al. aimed to estimate intersection topology and geometry as well as localize other traffic using a set of hand-crafted image features. In contrast to our work, they used stereo imagery from a small-scale dataset of 113 manually labeled intersections. In addition, their probabilistic model reasoned over multi-frame videos leading up to each intersection, while our model performs inference using a single monocular image as input. Yao et al. recently utilized a structured SVM coupled with edge, color, and homography features to segment the collision-free space in monocular video \cite{collisfree}. In our work, the set of driveable path headings (without regard to traffic) is one of several road attributes our model learns to estimate from a single frame.

A lidar-based approach to intersection detection was explored in \cite{lidar_inter}. Here, we focus on RGB imagery from cameras, bypassing the use of lidar which is both much more expensive and of lower resolution. While \cite{lidar_inter} focuses solely on identifying two types of intersection topology, our method focuses on inference about both intersections and regular road segments. In addition, our work focuses on a set of nine categorical and numerical road layout attributes, several of which have not been previously been studied in the context of computer vision.

\textbf{Deep learning for autonomous driving:}
In an early application of deep learning to autonomous driving \cite{muller}, the small, off-road, remote-controlled truck DAVE learned to avoid obstacles. A ConvNet was trained to map stereo image frames directly to an ideal steering angle for DAVE. In a similar system also trained in an end-to-end fashion, a group at Nvidia recently demoed a real car with steering controlled by a ConvNet \cite{nvidia_end2end}. In \cite{deep_driving}, Chen et al. trained a ConvNet to predict several "affordance indicators" (e.g., distance to the preceding car) for driving in a race car simulator. This is similar to our work in that we do not directly compute a driving action, but rather several road layout attributes that can later inform a driving controller. In \cite{deep_highway}, Huval et al. trained a ConvNet for car and lane detection on highway driving videos. In contrast to our image dataset, theirs was manually annotated via Amazon Mechanical Turk. While OpenStreetMap is also developed through crowdsourcing, in our system the ground truth label transfer to Google Street View images is fully automated.

\textbf{Leveraging OSM/GSV:}
Recently, researchers have exploited the OSM or GSV datasets for a few localization-related applications. Floros et al. \cite{openstreetslam} paired visual odometry with OSM to alleviate visual odometry's tendency to drift. In \cite{ma_sun}, models estimating road attributes including intersection presence and road type (trained with OSM-extracted labels) were used to speed up self-localization on driving video. In \cite{road_maps}, M\'attyus et al. developed a model to segment OSM roads in aerial images, thereby providing road widths and enhancing the map's accuracy and applicability to localization. Lin et al. \cite{ground2aerial} trained a ConvNet-based model to match GSV images with their corresponding location in aerial images. While we construct and leverage a dataset based on OSM and GSV, our model does not attempt to localize the street view image and instead estimates road layout attributes given the surrounding visual scene.

\section{Dataset}
Given a street scene panorama taken by a car-mounted camera, we wish to train a model to predict a set of road layout attributes. To train such a model, we first construct an appropriate large-scale dataset.

\textbf{Image collection:}
Google Street View contains panoramic images of street scenes covering 5 million miles of road across 3,000 cities. Each panorama has a corresponding metadata file storing the panorama's unique "pano\_id", geographic location, azimuth orientation, and the pano\_ids of adjacent panoramas. Beginning from an initial seed panorama, we collect street view images by running a bread-first search, downloading each image and its associated metadata along the way. Thus far, our dataset contains one million GSV panoramas from the San Francisco Bay Area. GSV panoramas can be downloaded at several different resolutions (marked as "zoom levels"). Finding the higher zoom levels unnecessary for our purposes, we elected to download at a zoom level of 1, where each panorama has a size of $832 \times 416$ pixels.  

\textbf{Map data:}
OpenStreetMap is an open-source mapping project covering over 21 million miles of road. Unlike proprietary maps, the underlying road coordinates and metadata are freely available for download. Accuracy and overlap with Google Maps is very high, though some inevitable noise is present as information is contributed by individual volunteers or automatically extracted from users' GPS trajectories. For example, roads in smaller cities may lack detailed annotations (e.g., the number of lanes may be unmarked). These inconsistencies result in varying-sized subsets of the data being applicable for different attributes. 

\textbf{Label transfer:}
Given latitude and longitude bounds for a set of street view images, we export the corresponding bounding box from OSM. Each road in OSM is represented with a a series of segments called "ways", each of which is made up of a series of nodes (a polyline) with geographic coordinates. In order to assign to each image a set of ground truth attributes, we must first match each image with the road segment it was captured on so that the road segment's annotations may be transferred over. 

We treat the panorama locations and way coordinates as points on a 2D plane\footnote{This approximation is acceptable because the region of focus here is
sufficiently distant from the poles.} (i.e., a Mercator projection) and compute the distance from each point to each polyline. For each panorama, the road segment found to be closest is selected. We note that this 2D method of matching images to streets is imperfect, as a GSV image on a bridge or street under a bridge may be incorrectly assigned. While this does add some noise to our training data, it occurs very infrequently.

In addition to outdoor street scenes, GSV also contains images captured within buildings as well as non-street scenes such as beaches. To ensure that most indoor images and non-street scenes are removed from the downloaded pool of images, we threshold the minimum distances to the road segments. Thresholding at 10.5 m seems to be effective while maintaining a high recall.

\section{Target attributes} \label{sec:targets}
Here we describe the road scene attributes, automatically extracted from OSM, that will serve as targets for prediction.
All images labeled with ground truth attributes are cropped and unwarped from the GSV panoramas at size $227\times227$ and with a wide field of view of 100\degree{}.

\begin{figure}
\centering 
\includegraphics[width=3in]{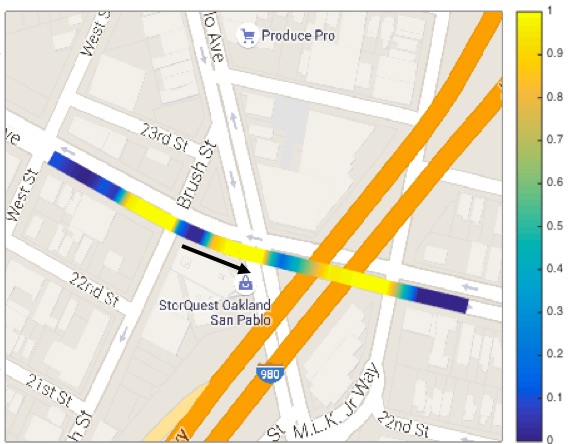}
\caption{Intersection detection heatmap. Images are cropped from test set GSV panoramas in the direction of travel indicated by the black arrow. The probabilities of "approaching" an intersection output by the trained ConvNet are overlaid on the road. (The images are from the ground level road, not the bridge.)}
\label{fig:inter_heatmap}
\end{figure}

\textbf{Intersections:}
Intersections are a source of dangerous driving scenarios, often accompanied by drastic changes in speed and complex interactions with other cars/pedestrians. Using our automated labeling procedure, we gather data to train our model to detect upcoming intersections (Figure \ref{fig:inter_heatmap}), predict the distance to them, and infer their topology. To this end, we first locate intersections on OSM by finding nodes shared by two or more road segments. Then, for panoramas located on the road segments involved in intersections, the distances and headings from the images to their nearest intersections are computed. 

\begin{figure}
\centering 
\includegraphics[width=3.45in]{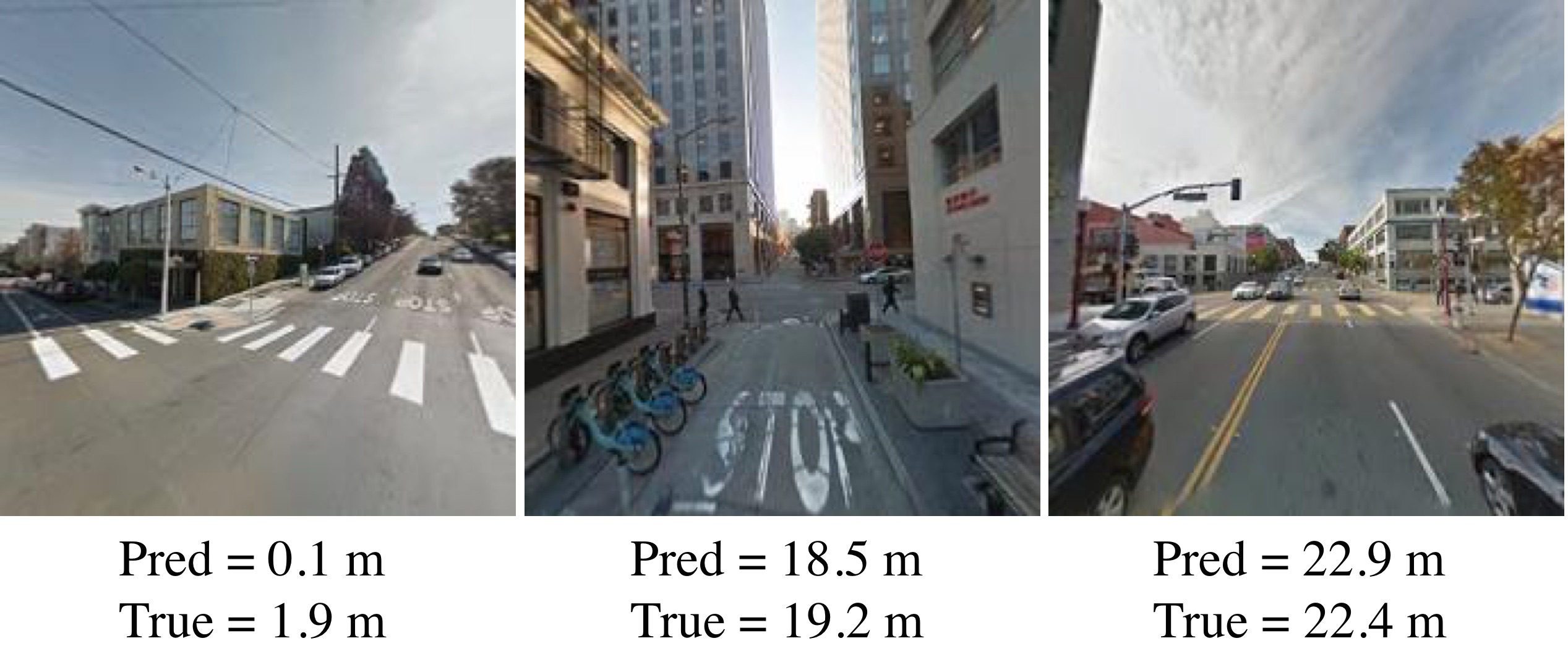}
\caption{Distance to intersection estimation. For images within 30 m of true intersections, our model is trained to estimate the distance from the host car to the center of the intersection across a variety of road types.}
\label{fig:inter_dists}
\vspace{-1ex}
\end{figure}

For the binary classification problem of there is/is not an intersection ahead, we face the ambiguity of not knowing if the closest intersection to a given GSV panorama is actually visible (no human checks the images during the labeling pipeline). After manually assessing a sample of images, we decided to count 30 m and below as positive and 100 m and above as negative. This thresholding provides a clear distinction between the two classes for ease of training. Images are cropped from the panoramas at the specified headings and labeled as positive (intersection) or negative (non-intersection). The distances to the intersections are also saved with the positive images to serve as targets for regression (Figure \ref{fig:inter_dists}). The topology of the intersections is addressed via "driveable headings" below. 

\begin{figure}
\centering 
\includegraphics[width=2.9in]{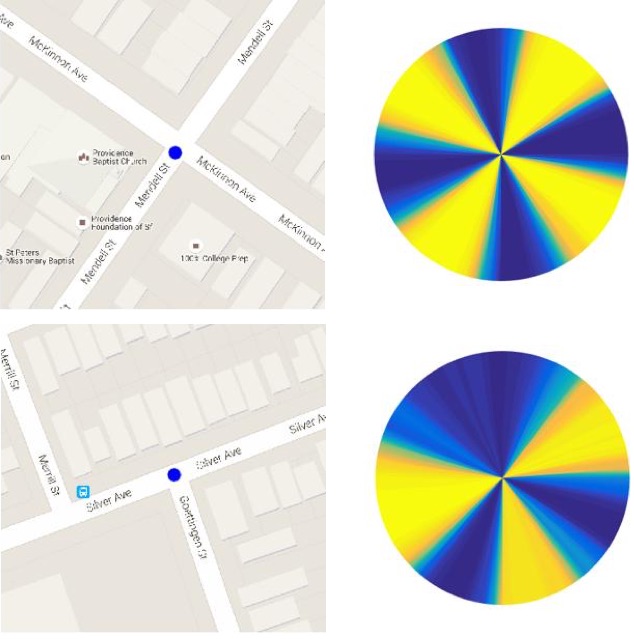}
\caption{Intersection topology is one of several attributes our model learns to infer from an input GSV panorama. The blue circles on the Google Maps extracts to the left show the locations of the input panoramas. The pie charts display the probabilities output by the trained ConvNet of each heading angle being on a driveable path (see Figure \ref{fig:inter_heatmap} for colormap legend).} 
\label{fig:interDrive}
\end{figure}

\begin{figure}
\centering
\includegraphics[width=3.45in]{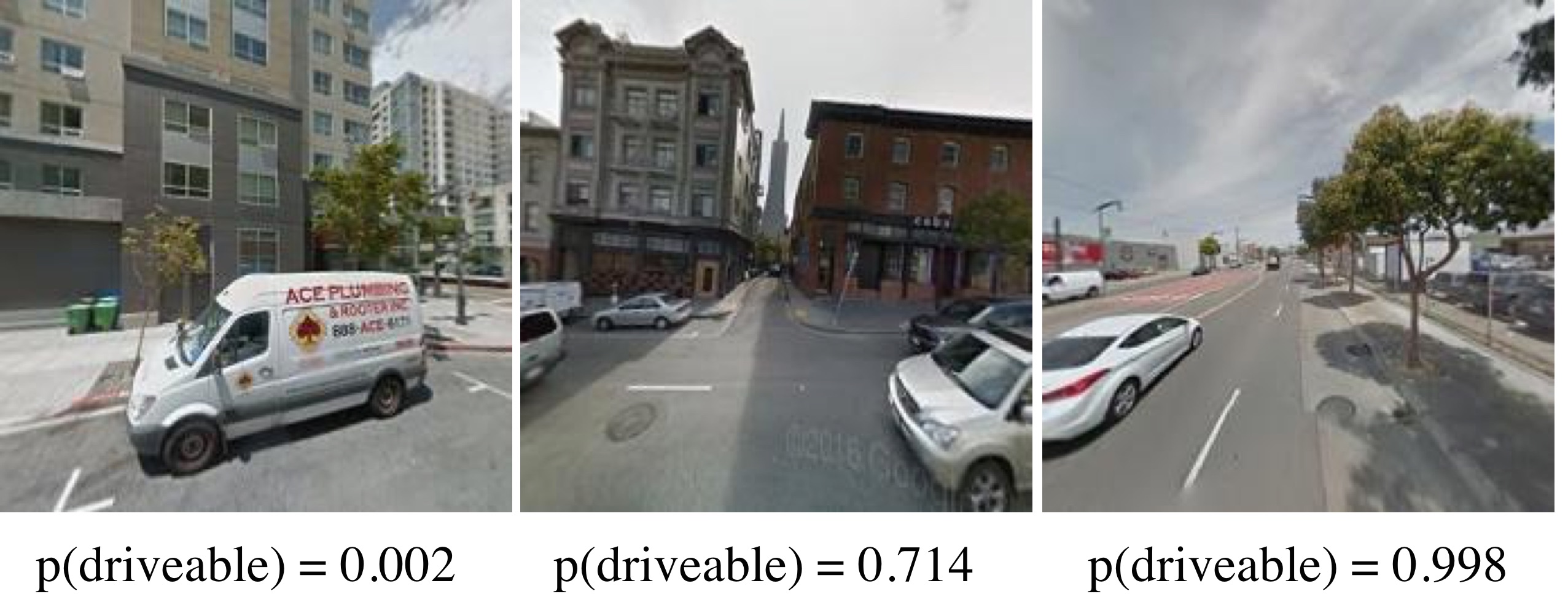}
\caption{Driveable headings. A ConvNet is trained to distinguish between non-drivable headings (left) and drivable headings aligned with the road (right). The ConvNet weakly classifies the middle example as drivable because the host car's heading is facing the alleyway between the buildings.}
\label{fig:driveability}
\end{figure}

\textbf{Driveable headings:}
The detection of driveable road paths (as opposed to sidewalk, buildings, etc.) is critical for corroborating portions of the HD maps used in autonomous driving. Given the specificity of annotations available here, the goal for our model will be to predict which headings (in the x direction) of a panorama align with driveable road. For our purposes, "driveable" simply means a street segment at the specified heading (with origin at the panorama's location) is present, disregarding specific lanes. Here, we define a heading to be driveable if it is within 22.5\degree{} of an actual road heading; other headings are considered non-driveable (Figure \ref{fig:driveability}). Using this setup, we can also label the driveable headings at intersections, allowing our model to learn to infer intersection topologies (Figure \ref{fig:interDrive}). 

\begin{figure}
\centering 
\includegraphics[width=3.45in]{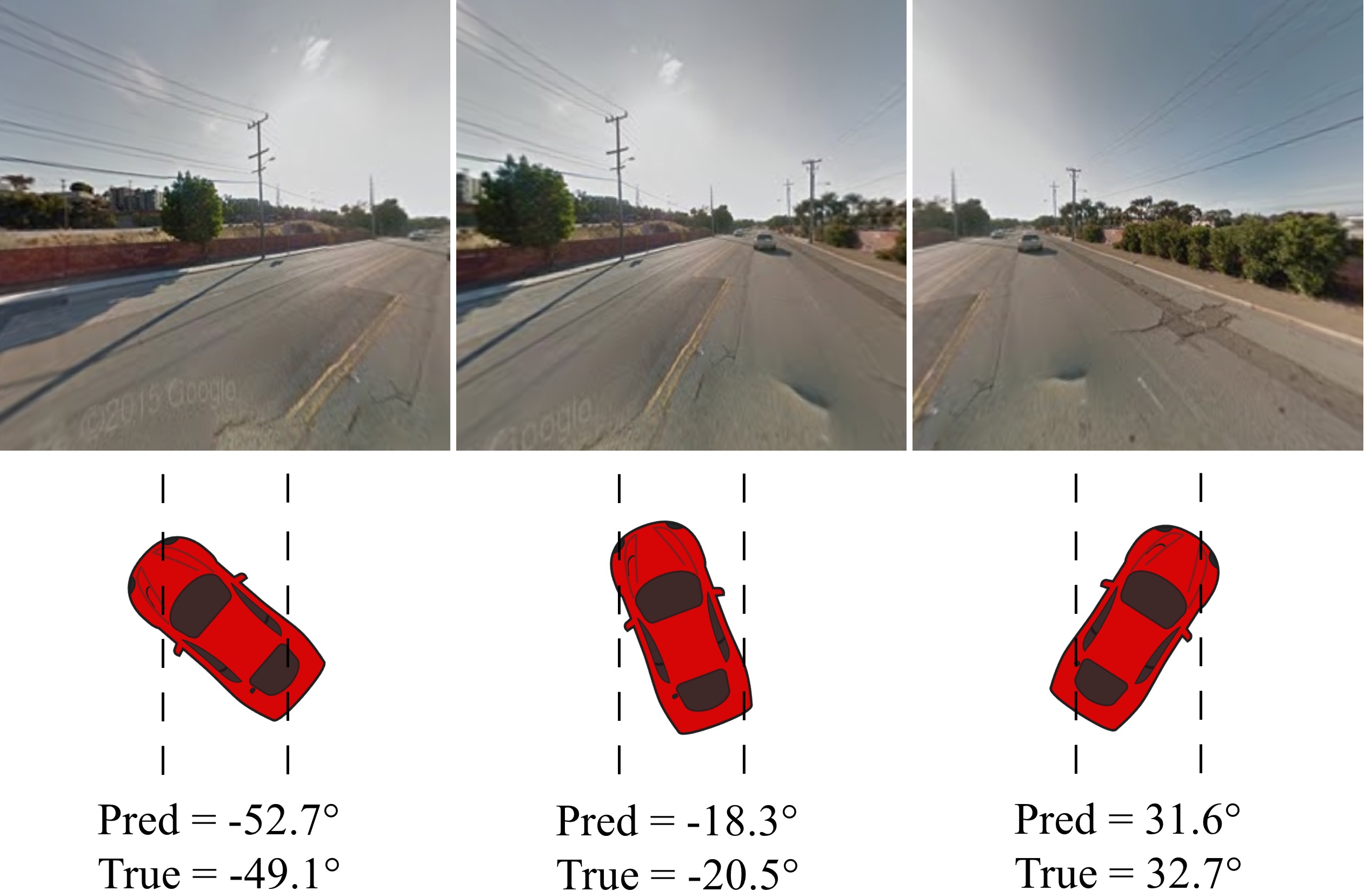}
\caption{Heading angle regression. The network learns to predict the relative angle between the street and host vehicle heading given a single image cropped from a GSV panorama. Below each GSV image, the graphic visualizes the ground truth heading angle.}
\label{fig:heading_reg}
\end{figure}

\textbf{Heading angle:}
While we treat driveable heading as a binary classification problem above, we can also treat this as a regression problem.
In this case, we crop images from each panorama up to 60\degree{} to the left and right of a true driveable heading and label them with the relative angle (Figure \ref{fig:heading_reg}). For this attribute, we do not include panoramas located at intersections as it is unclear how to label images facing in between two road segments involved in the junction. We include only those images that are at least 30 m away from the closest intersection.

\begin{figure}
\centering 
\includegraphics[width=3.45in]{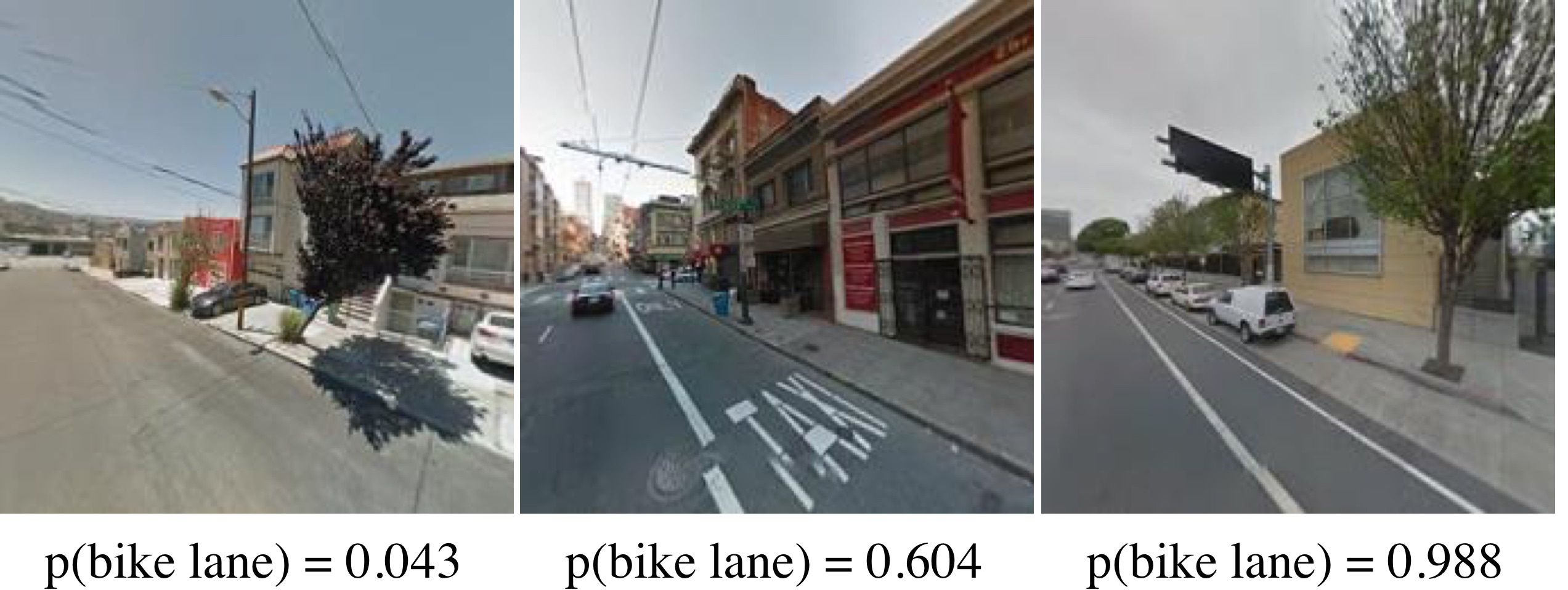}
\caption{The ConvNet learns to detect bike lanes adjacent to the vehicle. The GSV images are arranged from left to right in increasing order of probability output by the ConvNet of a bike lane being present (ground truth labels from left to right are negative, negative, positive). The middle example contains a taxi lane, resulting in a weak false positive.}
\label{fig:bikepaths}
\end{figure}

\textbf{Bike lanes:}
Roadside bike lanes indicate the possible presence of nearby bikers. Autonomous cars must be extra cautious and stay out of bike lanes, just as human drivers should. OSM roads often have the presence or absence of bike lanes annotated. For a given GSV image's road, if the presence of a standard roadside bike lane is labeled on OSM, we crop the panorama at 45\degree{} to the right (for right-hand traffic cities) relative to the forward road heading so the bike lane (or absence thereof) is maximally in view. 
Bike lane detection can then be treated as a binary classification problem for our model (Figure \ref{fig:bikepaths}).

We note that occasionally, a "way" (road segment) on OSM may be labeled as having a bike lane, but this does not mean the bike lane continues along the entire length of the road segment. Thus, this is a source of occasional noise in our training labels, as our automated labeling pipeline cannot account for this.

\begin{figure}
\centering 
\includegraphics[width=3.45in]{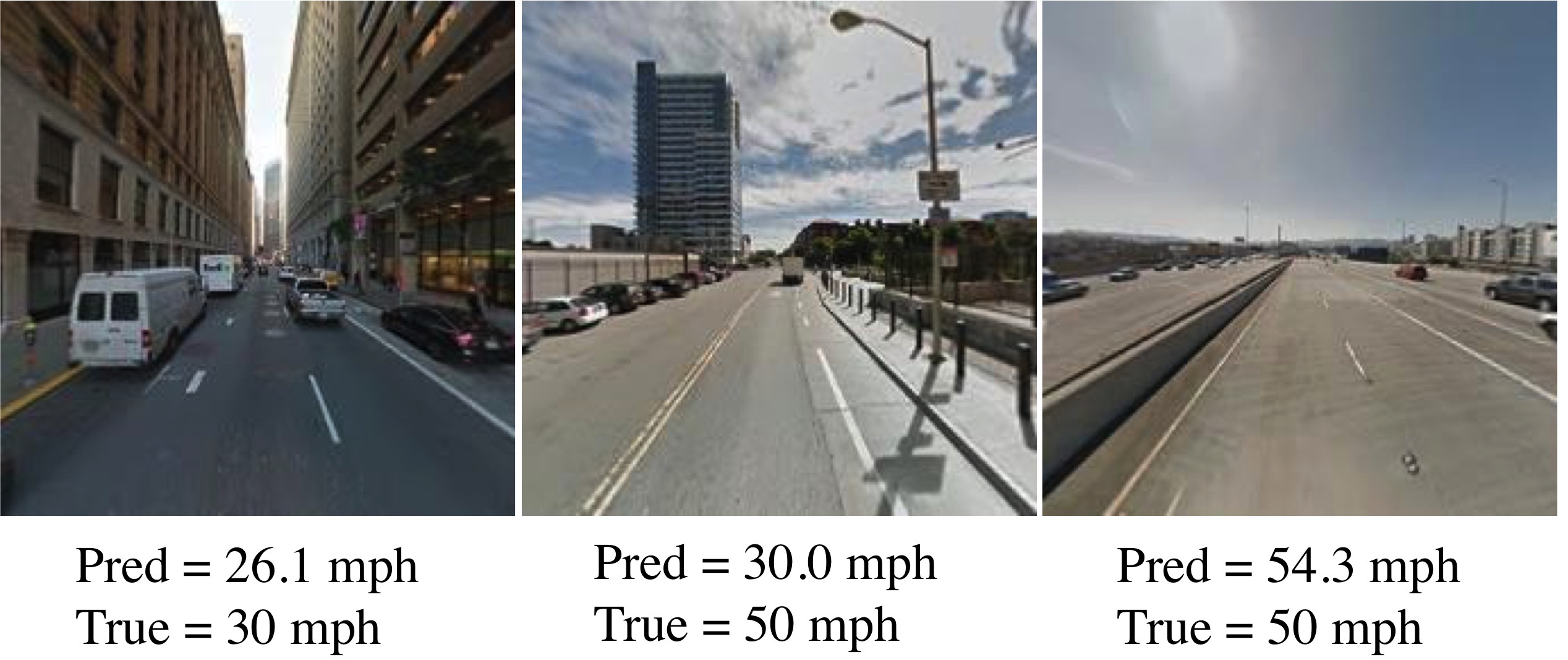}
\caption{Speed limit regression. The network learns to predict speed limits given a GSV image of road scene. The model significantly underestimates the speed limit in the middle example as this type of two-way road with a single lane in each direction would generally not have a speed limit as high as 50 mph.}
\label{fig:speed_lim}
\end{figure}

\textbf{Speed limit:}
As speed limits are key safety mechanisms for driving, it is important that an autonomous vehicle comes equipped with an intuitive notion of an appropriate speed given the surrounding environment. Obviously the road type, e.g., highway vs. small residential road, serves as a key indicator for this. Many OSM roads are labeled with speed limits, and these can trivially transferred over to GSV panoramas located on these roads. To collect a training set for this problem, we crop road-aligned images from the panoramas and label them with the corresponding speed limits (Figure \ref{fig:speed_lim}).

\begin{figure}
\centering 
\includegraphics[width=3.45in]{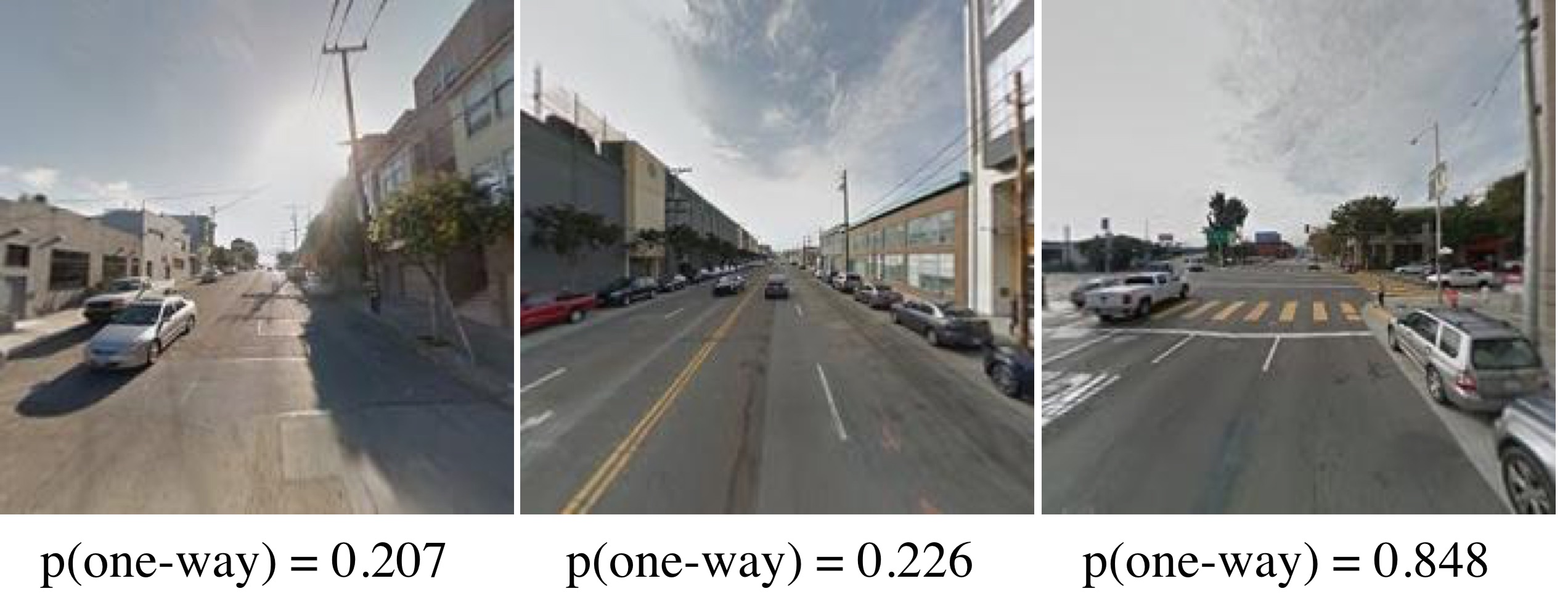}
\caption{One-way vs. two-way road classification. The probability output by the ConvNet of each GSV scene being on a one-way road is shown. From left to right the ground truth labels are two-way, two-way, and one-way. The image on the left is correctly classified as two-way despite the absence of the signature double yellow lines.}
\label{fig:oneway}
\end{figure}

\textbf{One-way vs. two-way:}
Knowing that a particular road is two-way allows an autonomous vehicle to anticipate oncoming traffic and help prevent changing into an incorrect lane. It is important that these cars can use on-board cameras to verify this type of information. Similarly to the previous section, one-way and two-way labels for roads on OSM can be trivially transferred over to corresponding GSV images to serve as training labels (Figure \ref{fig:oneway}).

\begin{figure}
\centering 
\includegraphics[width=3.45in]{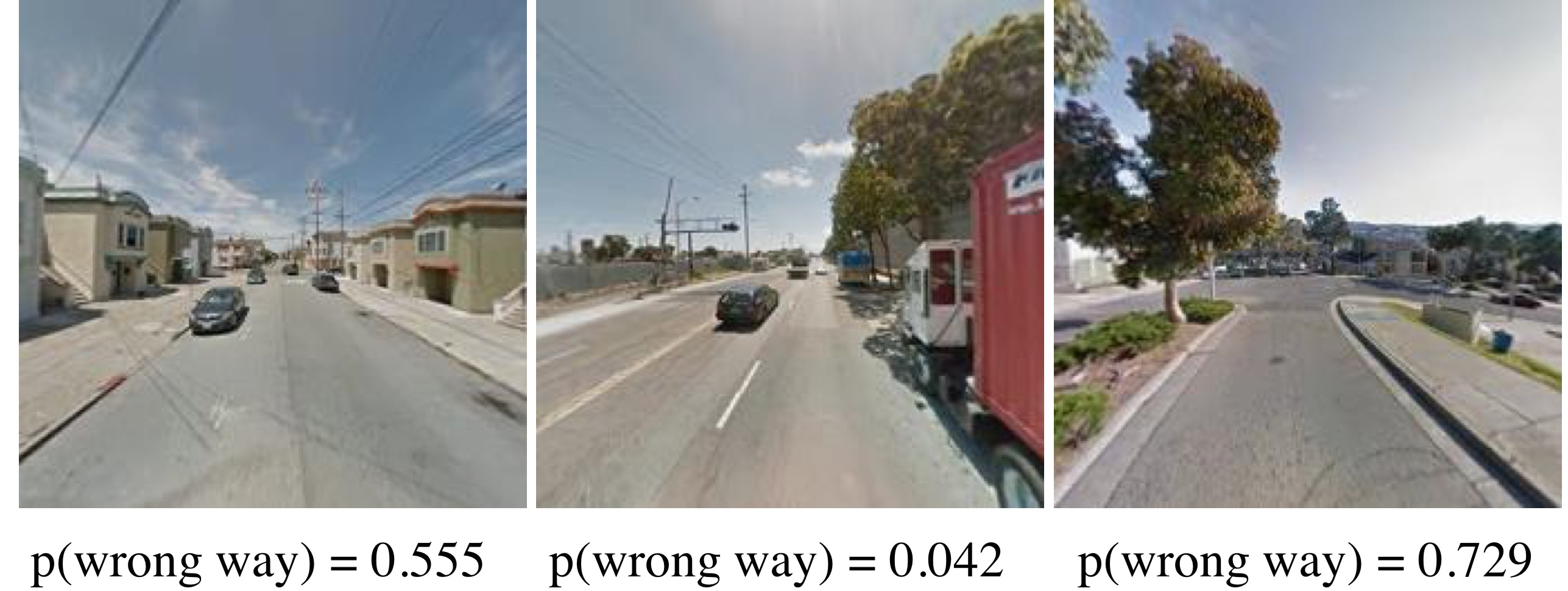}
\caption{Wrong way detection. The probability output by the ConvNet of each GSV image facing the wrong way on the road is displayed. From left to right the ground truth labels are wrong way, right way, and right way. For two-way roads with no lane markings (left), this is an especially difficult problem as it amounts to estimating the horizontal position of the host car. The problem can also be quite ill-defined if there are no context clues as is the case with the rightmost image.}
\label{fig:wrongway}
\end{figure}

\textbf{Wrong way detection:}
Inferring that a vehicle is facing the wrong way based on a single image captured by a car-mounted camera is a particularly challenging problem, requiring fine-grained reasoning about context clues. For example, a human faced with this challenge may know to look at the orientation of other cars, the direction street signs are facing, etc. To train our model for this task, we define "right" way images to be facing within 22.5\degree{} of the forward road heading and "wrong" way images to be facing within 22.5\degree{} of the backwards road heading (Figure \ref{fig:wrongway}).

\begin{figure}
\centering 
\includegraphics[width=3.45in]{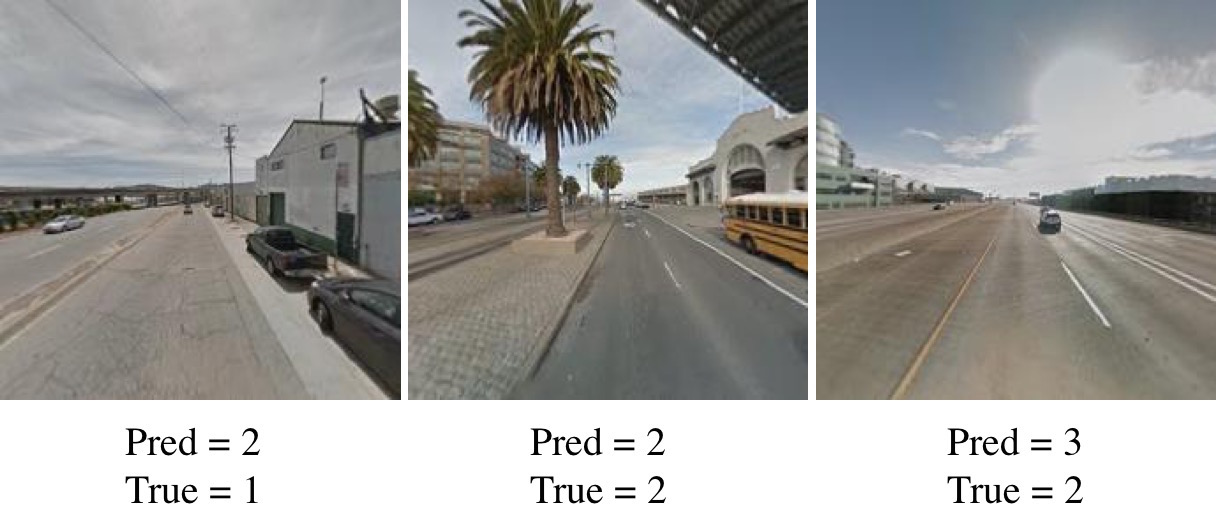}
\caption{Number of lanes estimation. The predicted and true number of lanes for three roads are displayed along with the corresponding GSV images. For streets without clearly visible lane markings (left), this is especially challenging. Although the ground truth for the rightmost image is two lanes, there is a third lane that merges just ahead.}
\label{fig:numlanes}
\end{figure}

\textbf{Number of lanes:} 
When planning for an upcoming turn/exit, the number of lanes on the road (and which lane the host car is in) determines what maneuvering will be necessary. The number of lanes is also often correlated with the speed limit. Here, we crop one forward image aligned with the road heading from each panorama and train our model to predict the number of lanes (Figure \ref{fig:numlanes}).
Unfortunately, there is substantial inconsistency in OSM regarding whether the number of lanes for two-way roads includes one or both directions of travel. To bypass this issue, we only include one-way roads in our training data for this attribute. 

\section{Model}

Our problem setup and labeling pipeline are agnostic to the actual ConvNet architecture employed. Here, we take the AlexNet architecture \cite{alexnet} and train it to predict both the categorical and numerical road layout attributes.

The target road layout attributes of focus in this work are related to each other, and there may be mutually useful image features for the different tasks. While this intuition suggests jointly training one network for every task, our experiments here use a separate network for each task. As the set of training images differs for each attribute, we focus on each attribute separately for simplicity. Of course, for deployment in a real vehicle, it would be preferable to have only a single network.

For the classification tasks and regression tasks, the networks are identical except for the loss layers. The classification networks have a standard softmax layer followed by a multinomial logistic loss layer. For the regression tasks, smooth L1 loss is used.

\section{Evaluation}

\textbf{Train-test split:}
To train our networks for road layout inference and subsequently evaluate their performance, we geographically split our dataset into distinct training and test sets. Specifically, we select a line of longitude such that 80\% of the GSV panoramas are to the west (training set) and the 20\% are to the east (test set). In this manner, the trained models are evaluated on previously unseen GSV imagery in novel locations. We note that a small portion of the training and test images may originate from the same road segments if they are very close to the longitudinal boundary. Our experimental results reported here are based on an initial iteration of our dataset of size 150K panoramas. 

\textbf{Network training:}
For each task, a network is trained with stochastic gradient descent using a batch size of 256 and learning rate of 0.001. To provide a good initialization for the ConvNets, we first pretrain them on the Places database \cite{places}. This dataset contains 2.5 million images across 205 scene categories, many of which are outdoors like the GSV imagery. Fine-tuning from this initialization greatly reduces the training time required. In our experiments, training tended to converge after a few thousand iterations for each task. The ConvNets are implemented in the Marvin deep learning framework \cite{marvin} and trained using a Tesla K40C. 

\textbf{Label balancing:}
The labels for a specific attribute's training data may be very unbalanced initially. For example, non-intersection GSV images are more common than those located at intersections. To counteract this imbalance in the case of classification tasks, random training instances from the sparser ground truth classes are duplicated until the classes are equally represented. Alternatively, the ConvNet data layer could be configured to randomly samples training instances for each batch uniformly among the classes. Similar balancing is carried out for the test set to allow for more interpretable performance results.

\textbf{Human baseline:}
To provide a reference point for the performance of our system on the newly collected dataset, we obtain a human baseline. We ask two volunteers to estimate the road layout attributes on 1000 randomly sampled images from each attribute's test set. The volunteers are each shown several example images and corresponding ground truth labels for each attribute prior to starting in order to clear up any ambiguity regarding attribute definitions.

While some of the categorical attributes are quite natural for a human to estimate given a single image (e.g., bike lane detection), some numerical attributes are not. For example, a human driver does not always think about distances to intersections or objects in numerical terms explicitly. Additionally, a human driver utilizes stereoscopic vision while we provide our volunteers with only monocular imagery. Nevertheless, we collect the human baseline performance for each attribute using the same monocular imagery that serves as input to the ConvNets. 

\textbf{Comparison to other methods:}
As both our dataset and target attributes are new, we do not quantitatively compare our approach to other methods. 

\textbf{Performance metrics:}
The performance for categorical attribute estimation is evaluated in percentage accuracy. 
For numerical attributes, we compute the mean absolute error (MAE). 

\begin{table}
\centering
\vspace{2ex}
\begin{tabular}{ l|c|c| }
			&								Human   &  Ours  \\ \hline
	Intersection Accuracy (\%, $\uparrow$)      		& \ 94.7		   & \ 88.3		\\ 
	Intersection Dist. MAE (m, $\downarrow$)			& \ 7.8		   & \ 4.3		   \\ 
	Driveable Accuracy (\%, $\uparrow$)				& \ 96.8		   & \ 94.9		   \\ 
	Heading Angle MAE (deg, $\downarrow$)			& \ 10.7		   & \ 9.2 	  \\ 
	Bike Lane Accuracy (\%, $\uparrow$)     			& \ 73.2		   & \ 75.9		   \\     
	Speet Limit MAE (mph, $\downarrow$)			& \ 11.6		   & \ 	6.8		\\ 
	One-way Accuracy (\%, $\uparrow$)    			& \ 75.1		   & \ 80.4		   \\ 	
	Wrong Way Accuracy (\%, $\uparrow$)                      & \ 82.2		   & \ 72.0		  \\ 
	Number of Lanes MAE ($\downarrow$)             		& \ 0.6		   & \ 0.9 
\end{tabular}
\caption{Accuracy of road attribute estimation. $\uparrow$ ($\downarrow$) indicates higher (lower) is better.} 
\label{table:results}
\vspace{-3ex}
\end{table}

\section{Results and Discussion}
\subsection{Road Layout Inference}
As shown in Table \ref{table:results}, the networks learn to estimate a variety of road layout attributes after training on the automatically labeled dataset. 

\textbf{Categorical attributes:} 
For three of the five classification tasks (detecting intersections, driveable headings, and wrong way orientations), the human baseline performs comparably to or better than the ConvNets. In particular, for wrong way detection, the human baseline achieves a 10 percentage point higher accuracy than the corresponding trained network. The background knowledge that humans can leverage for wrong way detection allows them to perform fine-grained reasoning about object orientations (e.g., street signs and other cars), while the model must learn to recognize these cues solely from examples. While on a two-way road with no visible lane markings, determining if the host car is facing the wrong way is especially challenging (Figure \ref{fig:wrongway}, left).

In both bike lane detection and one-way vs. two-way classification, the trained models achieve higher accuracies than the human baseline. In the case of bike lanes, the trained model is able to overcome the substantial noise inherent to this label (OSM does not indicate whether a bike lane continues along the entire length of a road segment) in addition to the bike lanes often being partially occluded. Surprisingly, the one-way vs. two-way classification ConvNet achieves a 6 percentage point higher accuracy than the human baseline, despite this task seeming to require fine-grained reasoning. For challenging scenarios when other vehicle orientations or lane markings are not visible for reference, the ConvNet may be learning to recognize some subtle cues about the general appearance of the road scene.

\textbf{Numerical attributes:}
For three of the four regression tasks (intersection distance, heading angle, and speed limit regression), the trained models perform comparably to or better than the human baselines. Impressively, the ConvNet trained to estimate distances to intersections is approximately twice as accurate as the corresponding human baseline. As mentioned previously, this result is not entirely unexpected as human drivers are not necessarily accustomed to perceiving the road layout in numerical terms. For example, when human drivers steer a vehicle, they have a general sense of their orientation relative to the road, but they do not need to know the exact angle in degrees. 

The network trained for number of lanes estimation did not perform well, achieving a MAE of 0.9 while the human baseline achieved 0.6. This task may be very ambiguous, especially when there are parking lanes on the side of the road or there are upcoming merging lanes (Figure \ref{fig:numlanes}, right). Additionally, the labeling of number of lanes is very sparse in OSM, with most roads not including this information. This factor, combined with our restriction to only one-way roads for this task, resulted in a substantially reduced training set of 12K GSV images.

\subsection{Application to Infrastructure Improvement}
Inspired by our experimental results, we propose that our method may be suitable for a novel application: improving city infrastructure. After being exposed to a large-scale pool of street view imagery and associated road attributes, the ConvNets are in essence learning what constitutes "normal" attributes for a given road of a certain visual appearance. Thus, for roads where the ground truth attribute set does not align well with its appearance, perhaps one or more of these attributes should be altered. 

For example, consider the center image in Figure \ref{fig:speed_lim}. The ConvNet has significantly underestimated the speed limit for this street, outputting a prediction of 30 mph when the ground truth is 50 mph. Is the trained ConvNet actually wrong? Perhaps the speed limit \textit{should} be 30 mph. After all, this is a two-way road with a single lane of traffic in each direction. If roads with such a setup are generally assigned lower speed limits, and this is what the ConvNet has learned, then the ConvNet's output may be interpreted as a recommended speed limit.

Similarly, roads that elicit a high probability from the trained model of being one-way when in fact they are two-way may be dangerous. If the network is fooled, it may be the case that the road/lane appearance does not imply two-way traffic strongly enough. Updating such roads accordingly may improve safety. With the capability of being deployed on a massive set of street view imagery automatically, these networks could quickly identify problem areas and recommend potential improvements.

We emphasize that this secondary application is only a proposal. Further work would be required to determine if such a system could effectively improve road safety.

\section{Conclusions and Future Directions}
In this work, we demonstrated that by leveraging existing street view image databases and online navigation maps, we can train models for road layout inference on a new, completely self-supervised dataset. Importantly, the automated labeling pipeline introduced here requires no human intervention, allowing it to scale with these large-scale databases and maps. Quantitative evaluation indicated that the trained convolutional networks learn to estimate a variety of attributes about the road layout given a single street view image as input, performing comparably to the human baseline on most tasks.

While we developed our model using street view images from one geographic region, it will be interesting to see how well the learned networks can transfer across distant regions (e.g., train on San Francisco, test on Paris). Perhaps only some fine-tuning will be necessary. Of course, certain characteristics of a given region, such as the traffic "handedness" (left-hand or right-hand traffic), may be taken into account to ensure that the cities grouped together have similar driving infrastructure.

As the tasks of focus here and the constructed dataset are both new, there is substantial opportunity for future improvements on our approach. For example, instead of using single frame convolutional networks, future work may examine the use of temporal models (e.g., recurrent networks) to estimate road layout attributes using contiguous sequences of street view images.

\addtolength{\textheight}{-12cm}   




\section*{ACKNOWLEDGMENTS}
 This work was supported in part by the Department of Defense through the NDSEG Fellowship Program.



\bibliographystyle{IEEEtran}
\bibliography{References}

\end{document}